# MPC-based Deep Reinforcement Learning Method for Space Robotic Control with Fuel Sloshing Mitigation

Mahya Ramezani, M. Amin Alandihallaj, Barış Can Yalçın, Miguel Angel Olivares Mendez, and Holger Voos

*Abstract—* This paper presents an integrated Reinforcement Learning (RL) and Model Predictive Control (MPC) framework for autonomous satellite docking with a partially filled fuel tank. Traditional docking control faces challenges due to fuel sloshing in microgravity, which induces unpredictable forces affecting stability. To address this, we integrate Proximal Policy Optimization (PPO) and Soft Actor-Critic (SAC) RL algorithms with MPC, leveraging MPC's predictive capabilities to accelerate RL training and improve control robustness. The proposed approach is validated through Zero-G Lab of SnT experiments for planar stabilization and high-fidelity numerical simulations for 6-DOF docking with fuel sloshing dynamics. Simulation results demonstrate that SAC-MPC achieves superior docking accuracy, higher success rates, and lower control effort, outperforming standalone RL and PPO-MPC methods. This study advances fuel-efficient and disturbance-resilient satellite docking, enhancing the feasibility of on-orbit refueling and servicing missions.

## I. Introduction

The limited fuel reserves of spacecraft and satellites constrain their operational lifespan and deep-space exploration. On-orbit refueling presents a transformative solution, extending mission duration, reducing costs, and improving adaptability. Additionally, refueling minimizes satellite launches, mitigates space debris, and supports satellite servicing, debris removal, and space infrastructure development, contributing to long-term space sustainability.

A major challenge in on-orbit refueling is ensuring precise and safe docking, as small velocity miscalculations can lead to mission failure. Fuel sloshing in microgravity introduces unpredictable forces, complicating spacecraft control.

Extensive research has explored sloshing dynamics through mass-spring and pendulum models [2], as well as Computational Fluid Dynamics (CFD) simulations. Studies have examined sloshing effects on spin-stabilized spacecraft [3], flexible satellites [4], and spacecraft trajectory control. Various control methods have been proposed, including fuzzy control [5-7], PD control [8, 9], LQR [10], H∞ control [11], sliding mode control [12, 13], and model predictive control (MPC) [14]. While these methods have advanced spacecraft attitude control, their application to fuel sloshing in docking scenarios remains underexplored.

MPC is a promising approach, capable of predicting system behavior and mitigating disturbances in real time. However, computational complexity and dependence on high accuracy sloshing models limit its real-time application.

In parallel, Artificial Intelligence (AI), particularly Reinforcement Learning (RL), has shown promise in aerospace applications, including trajectory [15] and path [16] planning in unknown environments [17], satellite collision avoidance [18], orbital maneuvering [19], satellite task scheduling [20, 21], motion control [22], and even failure predictions [23]. RL improves performance through iterative learning in simulated environments, reducing real-time computational demands. However, RL alone requires extensive training. Integrating RL with MPC can accelerate learning, leveraging MPC's predictive capabilities while benefiting from RL's adaptability to uncertainties [15].

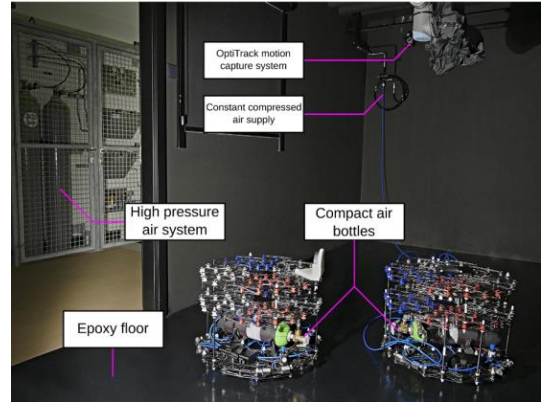

Figure 1. Two floating platforms operating in Zero-G Lab [1].

To address this, we propose a hybrid RL-MPC approach. Specifically, we incorporate Proximal Policy Optimization (PPO) with MPC principles. PPO ensures stable policy updates while maintaining adaptability, making it well-suited for initial docking control under uncertainty. By integrating MPC, PPO gains enhanced predictive capabilities, generating structured control sequences while adapting to environmental changes. This forms a baseline RL-MPC framework for satellite docking, tested in Zero-G Lab at the University of Luxembourg (shown in Fig.1). More information about the Zero-G Lab can be found in [24-28].

However, docking maneuvers, especially with fuel sloshing disturbances, require greater adaptability. Fuel sloshing introduces unpredictable forces affecting trajectory and attitude control. To tackle this, we enhance the PPO-MPC framework by incorporating Soft Actor-Critic (SAC), which maximizes entropy for better exploration and optimizes both policy and value functions efficiently. SAC improves sample efficiency and adaptability, crucial for real-time aerospace applications.

Transitioning from PPO-MPC to SAC-MPC, we develop a resilient docking control strategy that effectively mitigates fuel sloshing disturbances. This integrated methodology leverages MPC's predictive power and RL's adaptability, enabling precise docking in uncertain, dynamic conditions. Our study bridges RL-based control with practical aerospace applications, offering a robust solution for autonomous satellite docking with fuel-efficient, disturbance-resilient guidance.

The paper is structured as follows: Section 2 presents the orbital refueling mission's dynamic model. Section 3 and 4 detail the RL methodology and proposed framework. Section 5 validates the control strategy through numerical simulations, accounting for fuel sloshing dynamics. Section 6 concludes with key contributions and implications for on-orbit refueling missions.

## II. SYSTEM DESCRIPTION

Fuel sloshing, driven by the oscillatory motion of liquid, alters the center of mass and induces system-wide oscillations, affecting spacecraft stability. Traditional mass-spring and pendulum models approximate sloshing but become inaccurate in microgravity, where liquids exhibit non-cohesive behavior and can fragment into multiple blobs. To address this, OpenFOAM [29] is used for high-fidelity CFD simulations, enabling precise modeling of fluid fragmentation and irregular motion, improving control strategy reliability.

The mission scenario involves a tanker satellite performing docking maneuvers with a target satellite in Low Earth Orbit (LEO). While the system operates in six degrees of freedom (6-DOF), this study focuses on planar motion due to constraints in the Zero-G Lab at the University of Luxembourg (Fig. 1), where only 3-DOF motion (planar and rotational) can be replicated.

The system consists of a rigid tanker spacecraft with a spherical fuel tank (Fig. 3). Its dry mass (M) represents structural components, while fuel mass (m) is positioned b units from the spacecraft's center of mass. For attitude control, the tanker is equipped with 12 on/off thrusters. In the Zero-G Lab, the lab floor acts as an orbital plane analog, but due to 3-DOF limitations, the floating platforms use 8 thrusters instead of 12.

When considering sloshing as a disturbance and assuming the target satellite is in a circular orbit, the Clohessy-Wiltshire (CW) equations can describe the simplified relative orbital motion model as follows

$$\ddot{x} = 3n^2 x + 2n\dot{y} + \frac{F_x}{M} + d_{s_x}$$
$$\ddot{y} = -2n\dot{x} + \frac{F_y}{M} + d_{s_y} \quad (1)$$
$$\ddot{z} = -n^2 z + \frac{F_z}{M} + d_{s_z}$$

where $n$ is the orbital rate of the target satellite, $\delta r = [x \ y \ z]^T$ represents the tanker position with respect to the target satellite expressed in the RSW coordinate frame, $F_c = [F_x \ F_y \ F_z]^T = C_B^R [f_x \ f_y \ f_z]$ is the thruster force vector in the RSW frame, $C_B^R$ is the body frame to RSW frame rotation matrix, and $d_s = [d_{s_x} \ d_{s_y} \ d_{s_z}]^T = C_B^R \frac{f_s}{M}$ represents the fuel acceleration vector acting on the tanker. The relative equations of motion are derived in the RSW reference frame of the target satellite [30].

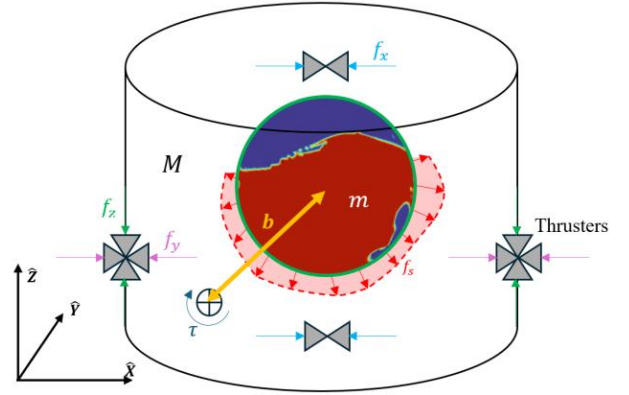

Figure 2. The tanker model with sloshing dynamics in the spherical tank.

The attitude dynamics and kinematics of the tanker spacecraft can be modeled as follows [31]

$$\mathbf{I}\dot{\boldsymbol{\omega}} = \boldsymbol{\omega} \times \mathbf{I}\boldsymbol{\omega} + \overbrace{\boldsymbol{\tau}_c + \boldsymbol{\tau}_s}^{\boldsymbol{\tau}}$$
$$\dot{\mathbf{q}} = \frac{1}{2}\begin{bmatrix} q_4 \boldsymbol{\omega} - \boldsymbol{\omega} \times \mathbf{q} \\ -\boldsymbol{\omega}^T \mathbf{q} \end{bmatrix} \quad (2)$$

in which $\boldsymbol{\omega}$ is the angular velocity vector, $\mathbf{I}$ represent the moment of inertia of the body, $\mathbf{q} = [q \ q_4]$ is the quaternion vector, $\boldsymbol{\tau}_c$ is the controlling torque generated by thrusters, and $\boldsymbol{\tau}_s$ represents the torque generated by the fuel motion. Consider the undisturbed tanker's nonlinear equations of motion as

$$\dot{s}(t) = f(s(t), u(t)) \quad (3)$$

where $s = [\delta r \ \delta \dot{r} \ q \ \omega]^T$ denotes the system state vector, $u = F_c$ represents the control input, and $f(\cdot,\cdot): (\mathbb{R}^n, \mathbb{R}^m) \to \mathbb{R}^n$ denotes a nonlinear function of the system. $n$ and $m$ are the number of system states and control inputs, respectively.

Using the first order approximation of the system (3), the resulting Linear Time-Varying (LTV) system is as

$$\dot{\hat{s}}(t) = \mathbf{A}_t \hat{s}(t) + \mathbf{B}_t u(t) + d(t) \quad (4)$$

where $\mathbf{A}_t \in \mathbb{R}^{n \times n}$, $\mathbf{B}_t \in \mathbb{R}^{n \times m}$ and $d(t) \in \mathbb{R}^{n \times 1}$.

The LTV system (4) can be utilized over a finite time horizon to approximate the nonlinear system (3) in case the effect of the fuel sloshing is neglected. This model is used to generate an optimum trajectory in the next chapter using MPC method.

## III. MPC FRAMEWORK

As discussed, the control strategy must ensure satellite safety by incorporating collision avoidance while managing disturbances like sloshing, which can introduce undesirable velocities during docking. However, directly modeling sloshing within the control framework is challenging due to its complexity and reliance on CFD simulations. To address this, the control scheme generates an optimal nominal trajectory based on unperturbed dynamics, while the RL component manages sloshing disturbances separately.

In this study, MPC is formulated as a path-planning problem, ensuring the docking trajectory is optimal and collision-free. The optimization problem solved at each time step is defined as follows

$$\min_{\boldsymbol{u}(t)} \int_{t=t_0}^{t_f} \left( \|\hat{\boldsymbol{s}}(t) - \boldsymbol{s}_f\|_\Omega^2 + \|\boldsymbol{u}(t)\|_R^2 \right) dt$$
$$\text{s.t. } \dot{\hat{\boldsymbol{s}}}(t) = \boldsymbol{A}_t \hat{\boldsymbol{s}}(t) + \boldsymbol{B}_t \boldsymbol{u}(t) + \boldsymbol{d}(t) \quad (5)$$
$$\hat{\boldsymbol{s}}(t) \in \Xi$$
$$\boldsymbol{u}(t) \in \boldsymbol{v}$$
$$\hat{\boldsymbol{s}}(t_0) = \boldsymbol{s}(t_0)$$

where $t_f$ denotes the final docking time, $t_0$ denotes the current time instant, $\boldsymbol{s}_f$ represents the desired final docking states, respectively. The system is also subjected to state constraint, $\Xi \in \mathbb{R}^n$, and control input constraint, $\boldsymbol{v} \in \mathbb{R}^m$, that are convex polytope sets. Given that the optimization problem (5) is a convex quadratic programming problem, it can be efficiently solved using standard Quadratic Programming solvers [32].

MPC generates an optimal docking trajectory by minimizing a cost function that balances tracking accuracy and control effort, serving as the nominal reference for docking. To handle sloshing disturbances, the RL module enhances the framework by ensuring robust and adaptive performance under uncertainty. Together, MPC and RL create an integrated control system capable of meeting the complex demands of satellite docking.

## IV. RL METHODOLOGY

The RL framework models the agent-environment interaction as an MDP, formally defined as a tuple $M = (S, A, P, R, \gamma)$, where $S$ is the state space, representing all possible system configurations, $A$ is the action space, encompassing all feasible control inputs, $P(s_{t+1} \mid s_t, a_t)$ is the state transition probability, governing the likelihood of transitioning from state $s_t$ to $s_{t+1}$ under action $a_t$, $R(s_t, a_t)$ is the reward function, defining the immediate reward for executing action $a_t$ in state $s_t$, and $\gamma \in (0,1]$ is the discount factor, determining the importance of future rewards relative to immediate ones. The agent interacts with the environment by selecting actions according to a policy $\pi(a_t|s_t)$, which dictates the probability of choosing action $a_t$ given state $s_t$.

### A. Proximal Policy Optimization (PPO)

PPO leverages an actor-critic framework, where the actor represents the policy, and the critic evaluates actions through a value function. The state-value function $V_w^\pi(\boldsymbol{s}_t)$ is expressed as [33]
$$V_w^\pi(\boldsymbol{s}_t) = E_\pi(\sum_{k=t}^T \gamma^{k-t} r_k(\boldsymbol{s}_k, \boldsymbol{a}_k) \mid \boldsymbol{s}_t) \quad (6)$$
where $w$ parameterizes the value function. The advantage function $A_w^\pi(\boldsymbol{s}_t, \boldsymbol{a}_t)$ quantifies the quality of actions relative to the expected value:
$$A_w^\pi(\boldsymbol{s}_t, \boldsymbol{a}_t) = \left( \sum_{k=t}^T \gamma^{k-t} r_k(\boldsymbol{s}_k, \boldsymbol{a}_k) \right) - V_w^\pi(\boldsymbol{s}_t) \quad (7)$$

The PPO loss function incorporates a policy probability ratio $p_t(\boldsymbol{\alpha})$:
$$p_t(\boldsymbol{\alpha}) = \frac{\pi_\alpha(\boldsymbol{a}_t|\boldsymbol{s}_t)}{\hat{\pi}_\alpha(\boldsymbol{a}_t|\boldsymbol{s}_t)} \quad (8)$$

This ratio compares the probability of selecting an action before and after a policy update. The PPO objective function is defined as
$$\mathcal{L}(\boldsymbol{\alpha}) = E_{p(\tau)} \left[ \min \begin{pmatrix} p_t(\boldsymbol{\alpha}) A_w^\pi(\boldsymbol{s}_t, \boldsymbol{a}_t), \\ \text{clip}[p_t(\boldsymbol{\alpha}), \epsilon] A_w^\pi(\boldsymbol{s}_t, \boldsymbol{a}_t) \end{pmatrix} \right] \quad (9)$$

where the clipping function constrains updates to ensure stability.

To improve the state-value function $V_w^\pi(\boldsymbol{x}_t)$, PPO minimizes the mean squared error loss:
$$L(w) = \frac{1}{2} E_{p(\tau)} \left[ \left( V_w^\pi(\boldsymbol{s}_t) - \left[ \sum_{k=t}^T \gamma^{k-t} r(\boldsymbol{s}_t, \boldsymbol{a}_t) \right] \right)^2 \right] \quad (10)$$

The clipping mechanism limits policy updates, preventing divergence while maintaining simplicity. To ensure stability, neural network outputs are scaled based on the running mean and standard deviation of state data. The policy network outputs are further normalized so that values of ±1 correspond to the maximum and minimum allowable thrust or torque.

### B. SAC Network Architecture

Unlike PPO, which typically employs two neural networks (actor and critic), SAC utilizes five neural networks:
- Policy Network (Actor): Predicts the probability distribution of actions.
- Two Critic Networks: Estimate the action-value function $\boldsymbol{Q}_\phi(\boldsymbol{s}_t, \boldsymbol{a}_t)$ to mitigate overestimation bias.
- Two Target Critic Networks: Stabilize training by providing reliable target values.

This architecture introduces redundancy and stability, with the target networks helping to correct overestimation issues inherent in value-based methods. SAC is widely regarded for its stability, sample Efficiency, and robustness.

SAC replaces the state-value function with the action-value function, $\boldsymbol{Q}_\phi(\boldsymbol{s}_t, \boldsymbol{a}_t)$, which evaluates the expected return for a specific action. The objective function for the policy network is given by [34]:
$$J(\theta) = E_{\boldsymbol{s}_t \sim D} \left[ \min_{i=1,2} \boldsymbol{Q}_{\phi_i}(\boldsymbol{s}_t, \tilde{\boldsymbol{a}}_t) \chi \log \pi_\theta (\tilde{\boldsymbol{a}}_t \mid \boldsymbol{s}_t) \right] \quad (11)$$
where $D$ is the replay buffer storing past experiences, $\tilde{\boldsymbol{a}}_t$ is a sampled action from the policy $\pi_\theta$, and $\chi$ regulates the influence of the entropy term. This objective combines reward maximization with entropy regularization, ensuring a balance between exploration and exploitation.

The critic networks in SAC minimize the Temporal Difference (TD) error using the following loss function:
$$L(\phi_i) = E_{\boldsymbol{s}_t \sim D} \left[ \left( \boldsymbol{Q}_{\phi_i}(\boldsymbol{s}_t, \tilde{\boldsymbol{a}}_t) - \boldsymbol{y}_t \right)^2 \right] \quad (12)$$
where
$$y_t = r_t + \gamma \left( \min_{i=1,2} \boldsymbol{Q}'_{\phi_i}(\boldsymbol{s}_{t+1}, \tilde{\boldsymbol{a}}_{t+1}) \chi \log \pi_\theta (\tilde{\boldsymbol{a}}_{t+1} \mid \boldsymbol{s}_{t+1}) \right) \quad (13)$$

Here $y_t$ is the target value for the critic networks, $r_t$ is the immediate reward, $Q'_{\phi_i}$ are the target critic networks, and $\tilde{\boldsymbol{a}}_{t+1}$ is an action sampled from the policy at the next state. This formulation ensures stability in training by minimizing errors between predicted and actual returns. The target critic networks are updated using a soft update mechanism:
$$\phi'_i = \omega \phi'_i + (1-\omega)\phi_i \quad i=1,2 \quad (14)$$
where $\omega \in [0,1]$ is the weighting factor. This approach ensures a gradual update of the target networks, reducing oscillations and stabilizing learning. The architectures of the SAC and PPO are compared in Table I.

TABLE I. PPO VS SAC NETWORK ARCHITECTURE.

| | PPO |
|---|---|
| Actor Network | 2 hidden layers (128, 64 neurons), tanh activation, output: 3 neurons (3DoF) or 6 neurons (6DoF), linear activation |
| Critic Network | 3 hidden layers (128, 64, 8 neurons), tanh activation, output: 1 neuron, linear activation |
| | SAC |
| Critic Networks | 3 hidden layers (256, 128, 64 neurons), tanh activation, linear output (action means), Softplus for standard deviations |
| Critic Networks | Two critic networks with 3 hidden layers (256, 128, 64 neurons), tanh activation, linear output (scalar Q-values), uses target critic networks for stability |

## C. Reward function

RL effectiveness relies on a well-defined reward function that minimizes state tracking errors, control effort, and fuel sloshing while reinforcing stabilization. These objectives are weighted through design coefficients.

This study evaluates two RL configurations: Standalone RL (PPO, SAC) and Integrated RL-MPC. Each has a distinct reward structure. In standalone RL, the algorithm independently learns optimal docking trajectories and control policies. The reward function is defined as follows:

$$r(\boldsymbol{s}_t, \boldsymbol{a}_t) = -\|\boldsymbol{u}_t\|_P^2 + \frac{\Psi_1}{1+e^{k(t)\delta}} + \frac{\Psi_2}{1+e^{k(t)\sigma}} - \|\boldsymbol{f}_s\|_\Phi^2 - \|\boldsymbol{\tau}_s\|_\eta^2 \quad (15)$$

where the first term penalizes control effort by introducing a quadratic cost on the control inputs, encouraging energy-efficient maneuvers. The second and third terms represent terminal stabilization bonuses. These terms ensure that as the system approaches its stabilization state, the reward increases exponentially, reinforcing successful docking and stabilization. Here, $k(t) > 0$ is a monotonically increasing function that gradually narrows the reward range over time, and $\sigma$ and $\delta$ represent position and rotation angle stabilization errors, respectively. The parameters $\Psi_1, \Psi_2 > 0$ control the magnitude of the terminal bonuses. The last two terms penalize forces and torques induced by fuel sloshing. By limiting these quantities, the reward function encourages a stable state for the liquid fuel during the maneuver, thereby minimizing unwanted movements caused by sloshing.

The integrated RL-MPC configuration introduces the output of MPC into the reward function to accelerate the RL training process. By incorporating an optimal trajectory from MPC, the RL agent is guided toward effective stabilizing trajectories, significantly reducing the learning time. The reward function in this framework is expressed as:

$$r(\boldsymbol{s}_t, \boldsymbol{a}_t) = -\left\|\dot{\hat{\boldsymbol{s}}}_t - \dot{\boldsymbol{s}}_t\right\|_M^2 - \|\boldsymbol{u}_t\|_P^2 + \frac{\Psi_1}{1+e^{k(t)\delta}} + \frac{\Psi_2}{1+e^{k(t)\sigma}} - \|\boldsymbol{f}_s\|_\Phi^2 - \|\boldsymbol{\tau}_s\|_\eta^2 \quad (16)$$

where the first term of the reward function quantifies the quadratic weighted error between the RL agent's state derivatives and the MPC reference trajectory, enabling the RL agent to learn effectively from MPC outputs. By using the MPC trajectory as a reference, the RL agent receives clear reward signals across the state-space, ensuring efficient learning of a stabilizing and optimal control strategy.

While standalone RL independently learns optimal trajectories, it requires longer training. The RL-MPC framework (Fig. 4) significantly reduces training time by combining RL's adaptability with MPC's precision, making it more efficient for complex satellite docking. Both methods prioritize energy efficiency, stabilization, and sloshing control, enhancing docking reliability.

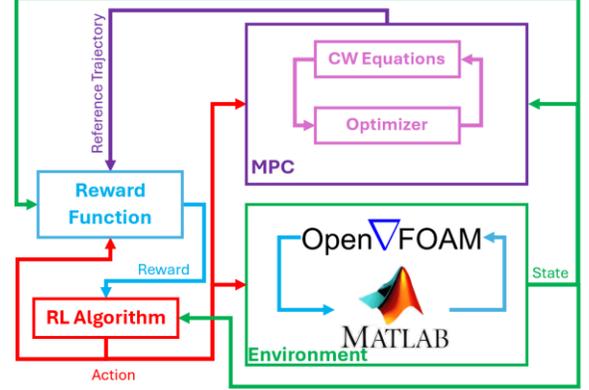

Figure 3. The schematic diagram of the integrated RL-MPC method.

## V. EXPERIMENT

The proposed MPC-RL integration is validated through two scenarios: a physical experiment in the Zero-G Lab and a numerical simulation of a 6-DoF satellite docking operation.

In the first scenario, experiments in the Zero-G Lab assess the impact of RL-MPC integration on planar stabilization. The floating platform setup provides a controlled environment to test stabilization under disturbances. Although fuel sloshing effects are not included due to experimental limitations, the results demonstrate that MPC enhances RL's convergence speed, robustness, and stabilization accuracy. The objective is to stabilize the platform at a fixed position.

The second scenario involves a high-fidelity numerical simulation of a 6-DoF satellite docking maneuver, incorporating fuel sloshing dynamics. Since these conditions cannot be replicated in the lab, four control methods—PPO, SAC, PPO-MPC, and SAC-MPC—are compared to evaluate stability and adaptability in real-world conditions.

By combining physical experiments and numerical simulations, this study offers a comprehensive evaluation of the proposed MPC-RL control strategies. The lab experiments validate the core concept in a simplified setup, while the simulations extend the analysis to complex space docking operations.

### A. Planar Stabilization in the Zero-G Lab

Experiments were conducted in the Zero-G Lab, which simulates near-frictionless motion using air-bearings. The goal was to stabilize a floating platform at a fixed point in the lab. The 5 m × 3 m × 2.3 m testing area features floating platforms actuated by eight nozzles, capable of generating 1 N of force per nozzle. Motion tracking is performed using an OptiTrack system with six Prime 13W cameras operating at 240 Hz, and control is integrated with ROS and MATLAB for efficient execution.

*1) Training Setup*

The RL algorithm, implemented in MATLAB, was trained in over 20,000 iterations to ensure policy convergence. Training data was collected in batches of 200 episodes, with force and torque commands generated every 0.1 seconds. Episodes were limited to 60 seconds for training and 100 seconds for testing, ensuring enough time for valid stabilization.

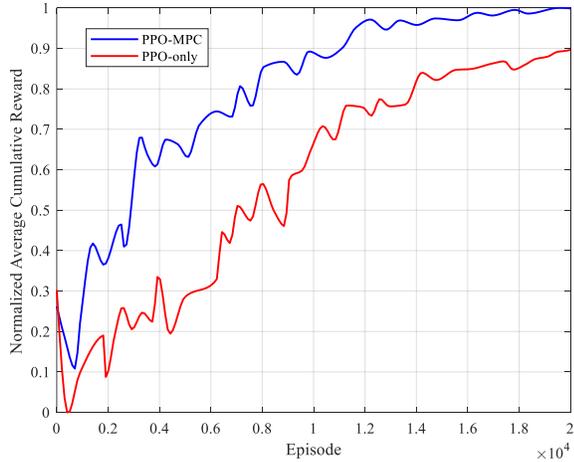

Figure 4. The average cumulative reward in the training phase.

The stabilization criteria are defined as position accuracy within 0.05 m, velocity within ±0.1 m/s, rotational accuracy within ±5 degrees, and angular velocity below ±1 degree per second.

The primary objective was to develop a robust feedback control policy capable of handling diverse initial conditions, ensuring adaptability to uncertainties. Training parameters are summarized in Table II, which are selected based on experience.

TABLE II. THE PPO TRAINING SETTING PARAMETERS.

| Parameter | Value | Parameter | Value |
|---|---|---|---|
| $KL_d$ | 0.001 | P | diag(10,10,10) |
| $\gamma$ | 0.98 | $\Psi_1$ | 100 |
| M | diag(1,10,5) | $\Psi_2$ | 10 |

*2) Training Results*

Fig. 5 illustrates the normalized average cumulative reward during training, showing that PPO-MPC consistently outperformed PPO-only. The PPO-MPC approach achieved faster convergence and higher rewards, demonstrating its efficiency and stability compared to PPO alone. This improvement is due to MPC providing optimal solutions that guide the RL agent's learning process, reducing the exploratory burden and accelerating policy optimization. In contrast, PPO-only relies solely on exploration, increasing the risk of getting trapped in suboptimal solutions. While PPO alone could eventually match PPO-MPC's performance, it would require significantly more training episodes.

*3) Experimental Results and Discussion*

After training, PPO-MPC and PPO-only were tested in real-world experiments using the floating platform in the Zero-G Lab. The platform was manually disturbed at four intervals, requiring the control algorithm to autonomously restore stabilization. These disturbances, marked by red arrows in the figures, assessed the adaptability and robustness of each method.

Fig. 6 illustrates the stabilization test sequence. Initially, the platform remains stable at the center, maintaining its position and orientation. Following a manual disturbance, it drifts away, reaching its maximum deviation before the control system activates to correct the trajectory and return it to its desired state. By the final snapshot, the platform is fully stabilized, demonstrating the control strategy's effectiveness in handling disturbances.

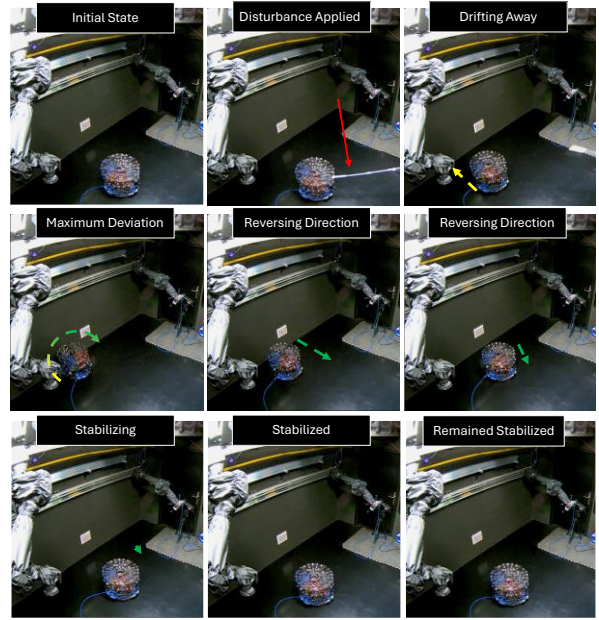

Figure 5. Time-lapse of the stabilization experiment in the Zero-G Lab.

Fig. 7 highlights PPO-MPC's performance, showing rapid stabilization after each disturbance, in which $dt = 0.1$ second. Even after a larger rotational deviation in the second disturbance, the platform quickly regained stability. Other disturbances primarily affected position, with minimal impact on orientation.

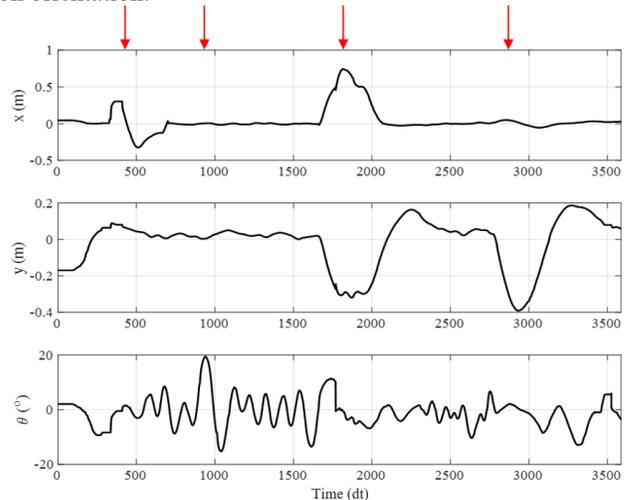

Figure 6. Performance of the PPO-MPC approach in disturbance rejection. Disturbances are shown by arrows.

Fig. 8 reveals PPO-only's limitations, with slower stabilization and larger errors. While the platform eventually returned to the origin, stabilization errors exceeded defined thresholds, with an average position error of 0.15m and a rotational error of 10 degrees. These findings align with the training phase, where PPO-only achieved lower rewards compared to PPO-MPC, confirming the benefits of MPC integration. More information on the control performance can be found in [35].

*B. Case 2: Docking Scenario*

A space simulation was conducted to assess the robustness and adaptability of PPO, SAC, PPO-MPC, and SAC-MPC in docking maneuvers. The scenario features a LEO satellite at

600 km altitude as the refueling target, with the docking tanker starting within a 200 × 120 m admissible zone in the RSW reference frame (Fig. 9). The blue marker represents the tanker, while the green marker denotes the target satellite.

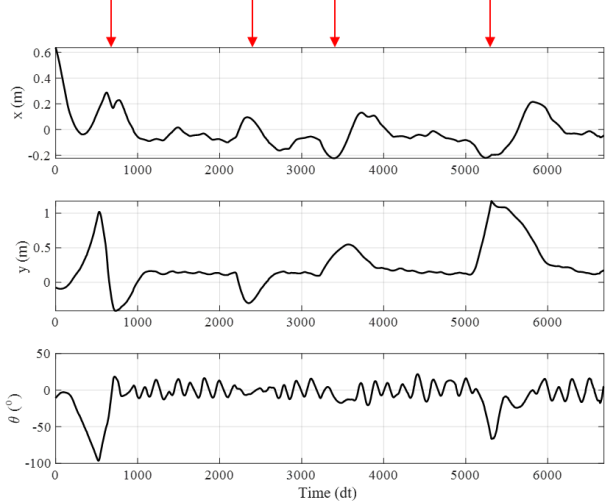

Figure 7. Performance of the PPO-only approach in disturbance rejection. Disturbances are shown by arrows.

To ensure safe docking, the tanker must stay at least 10 m from the target to allow rotational adjustments and remain within the admissible zone to uphold the CW equations. It is equipped with a partially filled spherical fuel tank (1 m radius) and thrusters producing up to 10 N of force (Table III).

Simulations were conducted under microgravity, incorporating fuel sloshing effects using the InterFoam solver, which applies the Volume of Fluid method to track the liquid-gas interface. The spherical tank (1 m radius) was discretized into 500,000 mesh cells for high-resolution modeling, which could provide an acceptable level of accuracy in the simulations. Boundary conditions included no-slip constraints at tank walls, velocity and pressure continuity at the liquid-gas interface, and adaptive time-stepping for numerical stability (Courant number ≤ 0.5).

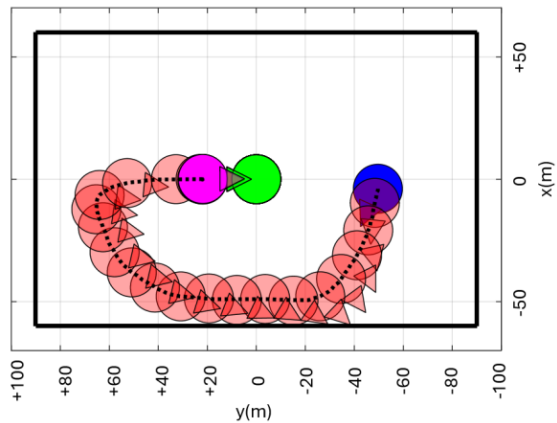

Figure 8. The docking admissible zone, initial position of the tanker, and the docking trajectory toward the target satellite.

The liquid phase properties were set to 900 kg/m³ density and 0.1 Pa·s viscosity, while the gas phase was treated as a vacuum. This high-fidelity simulation replicates microgravity docking conditions, enabling an accurate evaluation of RL-MPC control strategies for precise and stable docking.

TABLE III. THE TANKER'S MODEL PHYSICAL CHARACTERISTICS.

| Parameter | Value | Unit |
|---|---|---|
| $m$ | 50-300 | kg |
| $M$ | 250 | kg |
| $l$ | 1 | m |
| $b$ | 0.3 | m |
| $I$ | diag(100,100,150) | Kg·m² |

*1) Training Setup*

For training, the tanker's initial position and attitude were randomly selected within the admissible docking zone. The control algorithms were also tested under varying fuel levels, ranging from 10% to 90% of the tank's capacity in 20% intervals, to analyze how fuel sloshing and distribution affect performance.

The docking control system was implemented using MATLAB and OpenFOAM, with training conducted over 50,000 iterations to ensure RL network convergence. Training data was collected in batches of 100 episodes, with updates to the policy and state-value functions. Each training episode lasted up to 5000 seconds, balancing data efficiency and successful docking execution. For testing, the time limit was extended to 10,000 seconds to allow for complete docking trajectories. The training parameters are summarized in Table IV.

*C. Training Results*

Fig. 10 illustrates the fuel phase fraction inside the tank over four sequential frames, showing two different control scenarios captured at constant time intervals at the final stages. In the first row, the SAC-MPC method actively minimizes sloshing effects by incorporating them into the reward function, and in the second row, the MPC method is used without explicitly modeling or limiting sloshing-induced forces and torques.

TABLE IV. THE TRAINING SETTING PARAMETERS FOR CASE 2.

| Parameter | Value |
|---|---|
| $\gamma$ | 0.98 (PPO) / 0.99 (SAC) |
| $KL_d$ | 0.001 (PPO) |
| $\beta_\alpha$ | $1 \times 10^{-4}$ (PPO) / $1 \times 10^{-4}$ (SAC) |
| $\beta_w$ | $1 \times 10^{-4}$ (PPO) / $1 \times 10^{-4}$ (SAC) |
| Replay Buffer Size | $10^6$ (SAC) |
| Batch Size | 200 (PPO) / 256 (SAC) |
| $\tau$ | 0.005 (SAC) |
| P | diag(10,10,10,10,10,10) |
| $\eta$ | 10 |
| M | diag($1_{1\times3}, 10_{1\times3}, 5_{1\times3}, 50_{1\times3}$) |
| $\Psi_1$ | 100 |
| $\Psi_2$ | 10 |
| $\Phi$ | 5 |

In the MPC-only scenario, fuel sloshing remains significant throughout the tanker's movement. The effect becomes more pronounced during docking, where uncontrolled fuel motion induces forces and torques that could compromise stability in this critical phase. In contrast, the SAC-MPC method shows a clear reduction in fuel motion. While sloshing is initially present, it gradually diminishes, and by the final docking frames, the fuel is nearly stationary, exerting minimal disruptive forces on the tanker. This stabilized fuel distribution prevents sudden destabilizing movements, leading to a smoother and safer docking process.

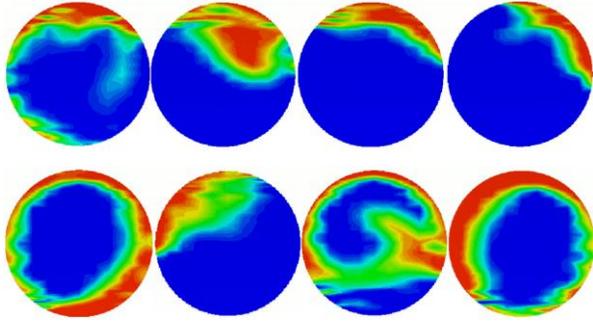

Figure 9. Fuel phase fraction inside the tank for SAC-MPC and MPC-only methods.

A Monte Carlo simulation was conducted to evaluate the four control configurations—PPO, SAC, PPO-MPC, and SAC-MPC under random initial conditions. The simulation analyzed key performance metrics, including success ratio, docking accuracy (position, velocity, and attitude errors), and control effort. Table V summarizes the results of over 1000 simulation runs for each method.

TABLE V. CONTROL METHOD PERFORMANCE METRICS

| Metric | PPO | SAC | PPO-MPC | SAC-MPC |
|---|---|---|---|---|
| Success Ratio (%) | 78 | 81 | 92 | 96 |
| Position Error (m) | 0.06 ± 0.03 | 0.05 ± 0.02 | 0.03 ± 0.01 | 0.02 ± 0.01 |
| Attitude Error (deg) | 0.5 ± 0.2 | 0.5 ± 0.15 | 0.3 ± 0.1 | 0.2 ± 0.08 |
| Control Effort (N·s) | 115 ± 15 | 114 ± 10 | 86 ± 8 | 79 ± 5 |

SAC-MPC achieved the highest success rate, followed by PPO-MPC, while standalone RL methods performed worse. The lower variance in SAC-MPC and PPO-MPC confirms greater reliability and consistency, as MPC integration provides optimal guidance, reducing RL's reliance on exploration. SAC-MPC also exhibited the best position accuracy outperforming PPO-MPC and significantly surpassing standalone RL methods, which had larger errors and higher variance. In attitude control, SAC-MPC achieved 0.2 ± 0.08 degrees error, while PPO-only had the highest errors, reflecting lower adaptability to disturbances. For control effort, SAC-MPC was the most efficient, followed by PPO-MPC. Standalone RL methods required significantly more thrust, highlighting SAC-MPC's optimized trajectory planning and energy efficiency.

Overall, SAC-based methods outperformed PPO-based ones, demonstrating higher success rates, better accuracy, and greater robustness to fuel sloshing disturbances. SAC-MPC consistently delivered the best performance, as MPC's predictive control reduced RL's reliance on exploration, improving docking precision, efficiency, and reliability.

## VI. CONCLUSION

This study presents an integrated RL-MPC framework for safe and precise satellite docking, addressing the challenges of fuel sloshing in microgravity. By combining MPC's predictive control with RL's adaptability, the proposed approach significantly improves convergence speed, stability, and fuel efficiency. Zero-G Lab experiments validate the method's effectiveness in a simplified setting, while 6-DOF simulations demonstrate robustness under real docking conditions.

Among all tested methods, SAC-MPC consistently outperformed others, achieving the highest docking success rate (96%), lowest position and attitude errors, and most efficient control effort. The integration of MPC into RL training accelerated learning and enhanced reliability by mitigating fuel sloshing disturbances. These findings support the application of RL-MPC-based control in on-orbit refueling, servicing, and autonomous docking missions, contributing to the long-term sustainability of space operations. Future work will explore hardware-in-the-loop testing and real-world spaceflight implementations.